\documentclass[runningheads]{llncs}
\usepackage{graphicx}

\usepackage{tikz}
\usepackage{comment}
\usepackage{amsmath,amssymb} 
\usepackage{color}

\usepackage{graphicx}
\usepackage{amsmath}
\usepackage{amssymb}
\usepackage{booktabs}

\usepackage{array} 
\usepackage{pifont} 
\newcommand{\cmark}{\ding{51}}%
\newcommand{\xmark}{\ding{55}}%

\usepackage[accsupp]{axessibility}  

\begin{document}
\pagestyle{headings}
\mainmatter
\def\ECCVSubNumber{4241}  

\title{MHR-Net: Multiple-Hypothesis Reconstruction of Non-Rigid Shapes from 2D Views} 


\titlerunning{MHR-Net}
%
\author{Haitian Zeng\inst{1, 2} \and
Xin Yu\inst{1} \and
Jiaxu Miao\inst{3} \and
Yi Yang\inst{3}}
\authorrunning{H. Zeng et al.}
%
\institute{University of Technology Sydney,\\ \and
Baidu Research,\\ \and
Zhejiang University,\\
\email{haitian.zeng@student.uts.edu.au,xin.yu@uts.edu.au}\\
\email{jiaxu.miao@yahoo.com,yangyics@zju.edu.cn}}
\maketitle

\begin{abstract}
We propose MHR-Net, a novel method for recovering Non-Rigid Shapes from Motion (NRSfM). MHR-Net aims to find a set of reasonable reconstructions for a 2D view, and it also selects the most likely reconstruction from the set.
To deal with the challenging unsupervised generation of non-rigid shapes, we develop a new Deterministic Basis and Stochastic Deformation scheme in MHR-Net. The non-rigid shape is first expressed as the sum of a coarse shape basis and a flexible shape deformation, then multiple hypotheses are generated with uncertainty modeling of the deformation part. MHR-Net is optimized with reprojection loss on the basis and the best hypothesis.
Furthermore, we design a new Procrustean Residual Loss, which reduces the rigid rotations between similar shapes and further improves the performance.
Experiments show that MHR-Net achieves state-of-the-art reconstruction accuracy on Human3.6M, SURREAL and 300-VW datasets.
\end{abstract}

\section{Introduction}

Recovering 3D structures from multiple 2D views is a classic and important task in computer vision. Non-Rigid Structure-from-Motion (NRSfM), which aims at reconstructing deformable shapes, is a challenging task and has been studied for decades.

The major difficulty of NRSfM is the ambiguity of solutions due to arbitrary deformation of shapes. Most of the NRSfM methods are based on the assumption of Bregler et al.~\cite{DBLP:conf/cvpr/BreglerHB00} where the deformable shape is a linear combination of a small number of atom shapes. This assumption greatly reduces the degree of freedom in NRSfM, yet it is still not enough for researchers to reach a deterministic and closed-form solution. Prior work of Akhter et al.~\cite{AkhterSK09} reveals that the local minimas grows exponentially with the basis number, and the reconstructed shapes from most of local minimas deviate significantly from ground truth. And Dai et al.~\cite{DaiLH14,DaiRank14} demonstrate that the rank minimization method also leads to multiple minimas in perspective cases. These reveal that there are usually multiple solutions that all minimize the cost function of a ``prior-free'' NRSfM, but it is generally intractable to find all those solutions due to the inherent complexity of NRSfM ambiguity.


Nevertheless, in most cases we only need to obtain several most reasonable hypotheses since they are valuable in practical scenarios, while searching for all ambiguous solutions exhaustively is not necessary. This leads us to focus on finding multiple high-quality hypotheses for the NRSfM problem.



Multiple hypotheses are usually modeled with uncertainty or generative models like CVAE~\cite{SohnLY15}, MDN~\cite{LiL19} or CGAN~\cite{LiL20}. However, these conventional modeling methods are supervised by 3D ground-truth, which is not available for a NRSfM problem. Moreover, in NRSfM, a naive ensemble of independent models is prone to decomposition ambiguity~\cite{DaiLH14}, and a variational autoencoder is also found hard to train~\cite{WangL21}.

To overcome the above challenges, we propose a novel MHR-Net for Multiple-Hypothesis Reconstruction of non-rigid shapes. Different from a standard model which outputs one reconstruction for a single input, MHR-Net is capable to produce multiple reasonable solutions and one best solution. We develop several critical designs for the successful generation of multiple hypotheses. Firstly, one non-rigid shape is expressed as the sum of a \emph{basis} and a \emph{deformation}. The basis is the coarse and shared structure among all shapes, while the deformation accounts for the diverse and flexible parts of shapes. This shape expression enhances the representation capability of MHR-Net when trained with an intermediate reprojection loss on the basis. 
Based on this expression, we further propose a novel Deterministic Basis and Stochastic Deformation (DBSD) scheme for multiple hypotheses generation. Specifically, MHR-Net estimates one basis in a standard deterministic manner and multiple deformations in a stochastic way. Then the multiple reconstructions are obtained by adding the basis and deformations. To optimize MHR-Net, we adopt a pseudo ``hindsight'' loss which is to select a hypothesis with the minimal reprojection error and calculates the standard loss function on the selected hypothesis. In inference, the model produces the best hypothesis in the same way. The DBSD scheme not only enables MHR-Net to produce multiple high-quality solutions of NRSfM, but also further enhances the accuracy of the reconstruction. 

Moreover, we develop a new Procrustean Residual Loss to regularize the reconstruction and reduce undesirable rigid rotations in a differentiable and efficient way. Experiments on Human3.6M, 300-VW and SURREAL datasets demonstrate state-of-the-art reconstruction accuracy of MHR-Net. Finally, we show that MHR-Net is capable to produce multiple possible solutions of 3D human poses and largely-deforming regions of dense human body meshes.

We summarize our contributions as follows:
\begin{itemize}
    \item We propose the novel MHR-Net for NRSfM. To the best of our knowledge, it is the first method that produces multiple high-quality hypotheses for non-rigid shape reconstruction in one model.
    \item We introduce a deterministic basis and stochastic deformation scheme together with a intermediate loss and a pseudo hindsight loss. These designs are effective for the challenging unsupervised uncertainty modeling of multiple 3D shapes. 
    \item We develop a novel Procrustean Residual Loss for NRSfM regularization, and it further improves the shape recovery accuracy of MHR-Net.
\end{itemize}

\begin{figure}[t]
\begin{center}
   \includegraphics[width=1.0\linewidth]{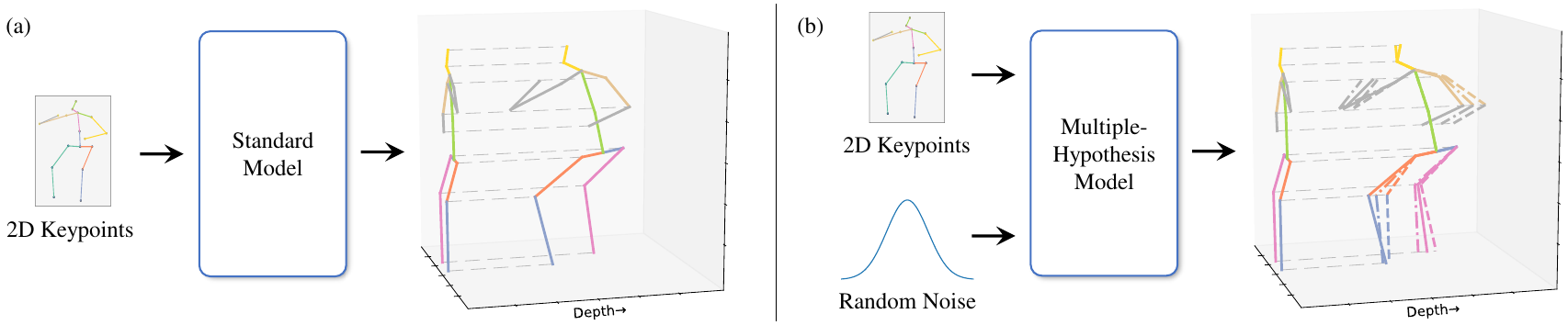}
\end{center}
   \caption{(a) A standard function-based NRSfM model maps the 2D input to a single 3D output. (b) The proposed MHR-Net is aware of the uncertainty of 2D-to-3D mapping. Given an extra noise vector, MHR-Net is capable to output multiple reasonable reconstructions of the 2D input.}
\label{fig:fig1}
\end{figure}

\section{Related Works}
\label{sec:intro}

\noindent
\textbf{NRSfM}. Recovering the deforming 3D shape from multiple 2D views is known as the Non-Rigid Structure-from-Motion problem (NRSfM), which is firstly put forward by Bregler et al.~\cite{DBLP:conf/cvpr/BreglerHB00}. NRSfM is a highly ambiguous problem, and therefore enforcing correct constraints is key to solving this problem. Previous works propose various effective approaches to the non-rigid shape reconstruction, including rank minimization~\cite{Dai2012A,Paladini2010SequentialNS}, smooth trajectories~\cite{5639016,5728827,DBLP:conf/nips/FragkiadakiSAM14,DBLP:journals/ijcv/Bue13}, manifold~\cite{6126319,DBLP:conf/cvpr/Kumar19}, metric projection~\cite{Paladini2009FactorizationFN}, sparsity-based methods~\cite{9099404}, energy minimization~\cite{DBLP:conf/cvpr/RussellFA11,DBLP:conf/eccv/SidhuTGAT20}, inextensibility~\cite{DBLP:conf/eccv/VicenteA12,DBLP:conf/cvpr/ChhatkuliPCB16}, isometry~\cite{DBLP:journals/pami/ParasharPB18}, deep models~\cite{9099404,Novotny2020C3DPO,DBLP:conf/eccv/ParkLK20,Zeng21}, procrustean normal distribution~\cite{7527684}, consensus~\cite{8778692}, hierarchical priors~\cite{4359359}, force-based and physic-based methods~\cite{AgudoM15,AgudoMCM16}, union of subspaces~\cite{6909596,DBLP:journals/corr/KumarDL17,DBLP:conf/cvpr/AgudoM17,Agudo2020Unsupervised3R,DBLP:conf/accv/AgudoM16}, piecewise methods~\cite{DBLP:conf/eccv/FayadAB10,Taylor2010NonrigidSF}, and many other breaking-through methods~\cite{DBLP:conf/cvpr/KumarCDL18,DBLP:journals/ivc/BueSA07,DBLP:conf/cvpr/RabaudB08,DBLP:conf/cvpr/AgudoACM14,DBLP:journals/pami/SalzmannPIF07,DBLP:conf/eccv/ProbstPCG18,DBLP:conf/cvpr/ParasharSF20,DBLP:conf/eccv/IglesiasOO20}.

Much attention has been paid to the uniqueness and determinacy of NRSfM. Xiao et al.~\cite{DBLP:journals/ijcv/XiaoCK06} show that selecting a set of frames as the basis can lead to a unique closed-form solution. Akhter et al.~\cite{AkhterSK09} argue that the orthogonal constraints of rotations is indeed sufficient for a unique solution except for a rigid rotation and the major difficulty lies in the optimization. 
Dai et al.~\cite{Dai2012A} propose a block-matrix rank-minimizing method and analyze whether their method leads to a unique solution or multiple solutions. Park et al.~\cite{ParkSMS10} provide a geometric analysis showing that the quality of sequential reconstruction is affected by the relative motion of a point and the camera, and propose a novel measure reconstructability to indicate the reconstruction quality. Valmadre et al.~\cite{DBLP:conf/eccv/ValmadreL10} propose a deterministic approach to 3D human pose recovery by using the rigid sub-structure of human body.

\noindent
\textbf{Multiple-Hypothesis 3D Pose Estimation}. The ambiguity of monocular 3D human pose estimation has been noticed early~\cite{DBLP:conf/cvpr/SminchisescuT01}. Li et al.~\cite{LiL19} use a mixture density network or a Conditional GAN~\cite{LiL20} to output a set of plausible 3D poses from a 2D input. Sharma et al.~\cite{SharmaVBSJ19} propose to solve the ill-posed 2D-to-3D lifting problem with CVAE~\cite{SohnLY15} and Ordinal Ranking. Wehrbein et al.~\cite{Wehrbein21} use Normalizing Flows to model the deterministic 3D-2D projection and solve the ambiguous inverse 2D-3D lifting problem. The major difference between our work and multiple-hypotheses 3D pose estimation is that our model is trained without 3D ground truth.

\section{Preliminary}
In the classic non-rigid structure-from-motion problem, given $N_f$ frames 2D observations $\{\mathrm{W}_i\}_{i=1}^{N_f}$ of a deformable object as input, we are interested in factorizing $\mathrm{W}_i \in \mathbb{R}^{2 \times N_p}$ into a camera matrix $\mathrm{M}_i \in \mathbb{R}^{2 \times 3}$ and a shape matrix $\mathrm{S}_i \in \mathbb{R}^{3 \times N_p}$ such that:
\begin{eqnarray}
  \mathrm{W}_i = \mathrm{M}_i \mathrm{S}_i.
  \label{eq:proj_eq}
\end{eqnarray}
Here, we suppose that $\mathrm{S}_i$ is centered at zero such that the translation term is cancelled, and $N_p$ stands for the number of points. $\mathrm{M}_i$ is the composition of a projection matrix $\mathrm{\Pi} \in \mathbb{R}^{2 \times 3}$ and a rotation matrix $\mathrm{R}_i \in \mathrm{SO(3)}$ so that $\mathrm{M}_i=\mathrm{\Pi} \mathrm{R}_i$. For orthographic projection, $\mathrm{\Pi}$ is simply $\begin{bmatrix} \mathrm{I}_2 & \mathbf{0} \\ \end{bmatrix}$. In this work, we suppose that the camera projection $\mathrm{\Pi}$ is known, allowing us to focus on the estimation of rotation $\mathrm{R}_i$.

In the recent progress of NRSfM~\cite{9099404,Novotny2020C3DPO,DBLP:conf/eccv/ParkLK20}, $\mathrm{M}_i$ and $\mathrm{S}_i$ are modeled as functions of the input $\mathrm{W}_i$. One typical paradigm~\cite{Novotny2020C3DPO,XuCL021} is to first extract features from $\mathrm{W}_i$ using a backbone network $\mathcal{H}(\mathrm{W}_i)$ like \cite{DBLP:conf/iccv/MartinezHRL17}, and then to estimate the $\mathrm{M}_i$ and $\mathrm{S}_i$ with different network branches $\mathcal{F}_0$ and $\mathcal{G}$ subsequent to $\mathcal{H}$:
\begin{eqnarray}
  \mathrm{S}_i = \mathcal{F}_0(\mathcal{H}(\mathrm{W}_i)), \quad \mathrm{M}_i = \mathcal{G}(\mathcal{H}(\mathrm{W}_i)).
\end{eqnarray}
Modeling the factorization as a function enables NRSfM methods to be optimized on large-scale datasets, and allows models to directly perform reconstruction on unseen data. To train such models, the cost function usually contains a data term and a regularization term, represented as:
\begin{eqnarray}
\mathcal{L} = \sum_{i=1}^{N_f}\mathcal{L}_{\mathrm{data}}(\mathrm{W}_i,\mathrm{M}_i,\mathrm{S}_i)+\mathcal{L}_{\mathrm{reg}}(\mathrm{M}_i,\mathrm{S}_i),
\label{eq:cost_fun1}
\end{eqnarray}
where the data term $\mathcal{L}_{\mathrm{data}}$ is usually the reprojection error $\left\| \mathrm{W}_i - \mathrm{M}_i\mathrm{S}_i \right\|$ and the regularization term is versatile.

\begin{figure}[t]
\begin{center}
\includegraphics[width=1.0\linewidth]{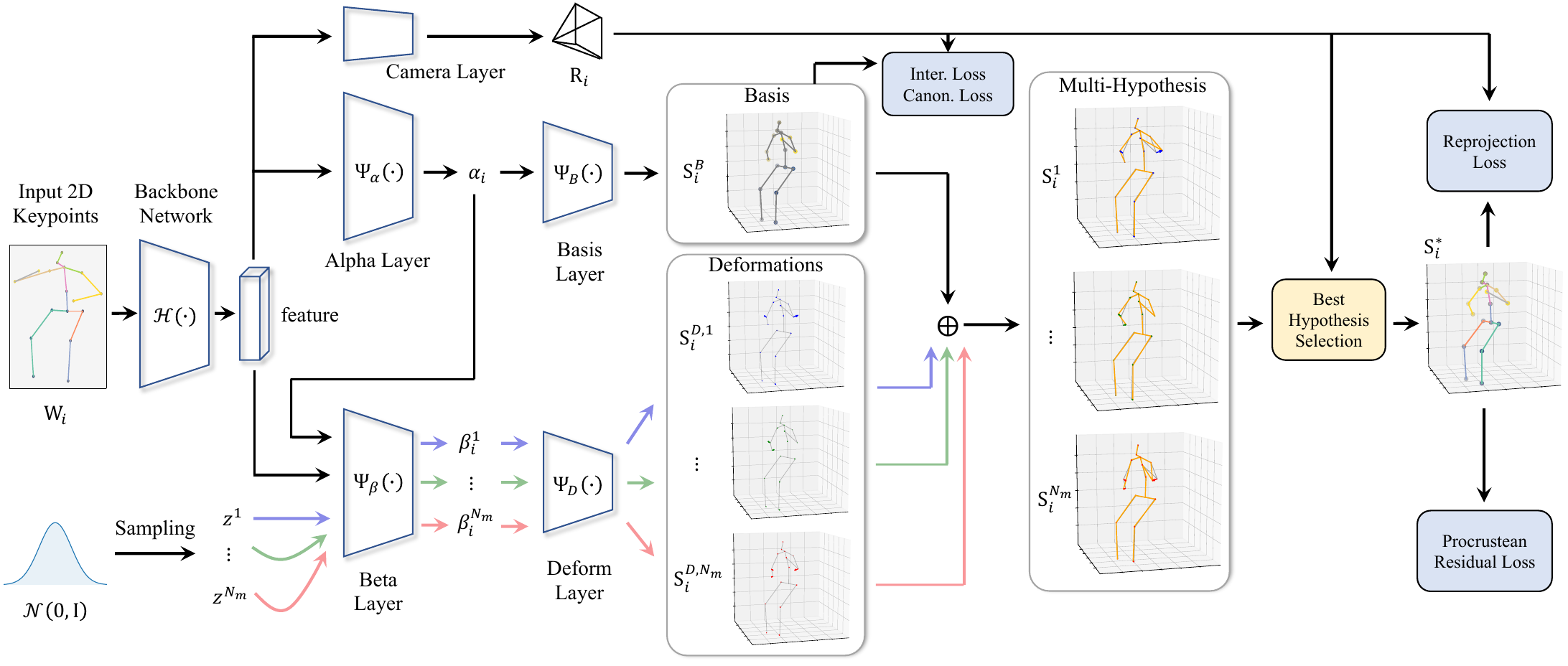}
\end{center}
\caption{An overview of the proposed MHR-Net. MHR-Net uses a backbone network $\mathcal{H}$ to extract features from the 2D input $\mathrm{W}_i$. The camera rotation $\mathrm{R}_i$ is estimated by the rotation layer. Next, the basis shape $\mathrm{S}_i^B$ and its coefficients $\alpha_i$ are estimated by $\Psi_B$ and $\Psi_\alpha$ respectively. To generate multiple hypotheses, beta layer $\Psi_\beta$ takes $\alpha_i$, $\mathcal{H}(\mathrm{W}_i)$ and random noise $\{z^1,\cdots,z^{N_m}\}$ as inputs, and outputs a set of deformation coefficients $\{\beta_i^1,\cdots,\beta_i^{N_m}\}$. Then the deformation coefficients are passed to the deformation layer $\Psi_D$ to produce deformations $\{\mathrm{S}_i^{D,1},\cdots,\mathrm{S}_i^{D,N_m}\}$. By adding each deformation to $\mathrm{S}_i^B$, multiple hypotheses $\{\mathrm{S}_i^{1},\cdots,\mathrm{S}_i^{N_m}\}$ are obtained. Then, the hypothesis with the smallest reprojection error is selected as the best hypothesis $\mathrm{S}_i^*$. Loss functions are calculated on $\mathrm{S}_i^B$ and $\mathrm{S}_i^*$.}
\label{fig:overview}
\end{figure}

\section{Proposed Method}
\label{sec:methods}
\subsection{Multiple Hypothesis Reconstruction Network - Overview}

We aim to develop a prediction function $\mathcal{F}$ that can output $N_m$ reconstructions for a single input $\mathrm{W}_i$:
\begin{eqnarray}
\mathcal{F}(\mathrm{W}_i)=\{\mathrm{S}^1_i,\mathrm{S}^2_i,\cdots,\mathrm{S}^{N_m}_i\},
\end{eqnarray}
and each of these reconstructions is supposed to minimize the cost function in Eq.~\ref{eq:cost_fun1}. As the inherent ambiguity of NRSfM is complex and there exists a large number of poor ambiguous solutions \cite{AkhterSK09}, we also expect the hypotheses to be as accurate as possible among all solutions.






However, generating multi-hypothesis reconstruction for NRSfM is challenging for several reasons: (1) Without 3D ground-truth as supervision, the ambiguous 2D-to-3D mappings cannot be learned using standard generative models like CVAE~\cite{SohnLY15}, Conditional GAN~\cite{LiL20} or Normalizing Flow~\cite{Wehrbein21}. (2) Multiple hypotheses easily suffer from the decomposition ambiguity of NRSfM~\cite{DaiLH14}, i.e. multiple solutions are trivial if they are related by a certain rotation $\mathrm{G}$ inserted in the decomposition $\mathrm{W}_i=\mathrm{M}_i\mathrm{G}\mathrm{G}^{-1}\mathrm{S}_i$. 

We introduce a novel Multiple Hypothesis Reconstruction Network (MHR-Net), which takes a step towards a multiple-hypothesis NRSfM model. MHR-Net overcomes the above difficulties in a simple and effective way, and is capable to produce multiple accurate reconstructions and one best hypothesis. Next, we describe the hypothesis generation scheme in Sec.~\ref{sec:dbsd}, the optimization in Sec.~\ref{sec:optimization} and the regularization in Sec.~\ref{sec:reg}.

\subsection{Deterministic Basis and Stochastic Deformation for Hypothesis Generation}
\label{sec:dbsd}

Traditionally, the 3D shape $\mathrm{S}_{i}$ is represented as a linear combination of $K_b$ atom shapes $\mathrm{B}_{k} \in \mathbb{R}^{3 \times N_p}$:
\begin{eqnarray}
  \mathrm{S}_{i} = \sum_{k=1}^{K_b} (\alpha_{i,k}\otimes \mathrm{I}_3)\mathrm{B}_k,
\end{eqnarray}
where $\alpha_{i,k}$ is the weight of $k$-th atom shapes in $\mathrm{S}_{i}$, and $\otimes$ is the Kronecker product. 
This widely-accepted representation implicitly assumes the low-rankness of all estimated 3D shapes, and it is one of keys to successful recovery of non-rigid shapes. However, the capacity of this shape representation is limited when applied to modern large-scale datasets since they usually contain millions of frames or thousands of keypoints. A naive way is to enlarge the capacity by increasing the dimension $K_b$, but it also bring undesirable degrees of freedom to the full shape, leading to a regularization-flexibility dillema.

Inspired by previous works~\cite{Agudo2020Unsupervised3R,DBLP:conf/cvpr/BartoliGCPOS08,DBLP:conf/eccv/SidhuTGAT20}, we develop a new shape representation for MHR-Net. We posit that the deformable shape is comprised of a \emph{basis} $\mathrm{S}_i^B$ and a \emph{deformation} $\mathrm{S}_i^D$, written as:
\begin{eqnarray}
  \mathrm{S}_{i} =
  \underbrace{\sum_{k=1}^{K_b} (\alpha_{i,k}\otimes \mathrm{I}_3)\mathrm{B}_k}_{\mathrm{S}_i^B} +
  \underbrace{\sum_{l=1}^{K_d} (\beta_{i,l}\otimes \mathrm{I}_3)\mathrm{D}_l}_{\mathrm{S}_i^D},
\label{eq:shape_expression}
\end{eqnarray}
where $\mathrm{D}_l \in \mathbb{R}^{3 \times N_p}$ denotes the $l$-th atom deformation, $\beta_{i,l}$ is the weight of $\mathrm{D}_l$ in $\mathrm{S}_{i}$, and $K_b < K_d$. Note that the mean shape for short sequences in \cite{DBLP:conf/eccv/SidhuTGAT20} is a special case of Eq.~\ref{eq:shape_expression} with $K_b=1, \alpha_i=1$. This basis-deformation modeling enhances the shape representation capability of MHR-Net with hierarchical flexibility. Intuitively, $\mathrm{S}_{i}^B$ is used to capture the low-rank common part of the 3D shapes, while $\mathrm{S}_{i}^D$ fits the diverse small deformations with higher flexibility. 

Based on the basis-deformation expression of shapes, we now introduce a novel Deterministic Basis and Stochastic Deformation (DBSD), which is the core scheme for multiple hypothesis generation in MHR-Net. We assume that the variation of $N_m$ accurate hypotheses appears only in $\mathrm{S}_{i}^D$.
While Wang et al.~\cite{WangL21} find that introducing randomness to the estimation of full 3D shapes with VAE~\cite{DBLP:journals/corr/KingmaW13} is not viable, the proposed partial uncertainty in $\mathrm{S}_{i}^D$ makes MHR-Net overcome the training difficulty. More importantly, we find that DBSD can lead to an even better recovery accuracy with the training strategy in Sec.~\ref{sec:optimization}. 


Specifically, MHR-Net generates multiple reconstructions by estimating one basis in the standard deterministic way and multiple deformations with stochastic variations. 
For the deterministic part, coefficients $\alpha_i=[\alpha_{i,1} \cdots \alpha_{i,K_b}]^\mathrm{T}$ are estimated with a convolutional layer $\mathrm{\Psi}_\alpha$:
\begin{eqnarray}
  \alpha_i = \mathrm{\Psi}_\alpha(\mathcal{H}(\mathrm{W}_i)).
\end{eqnarray}
For the stochastic deformation part, the coefficients $\beta_i=[\beta_{i,1} \cdots \beta_{i,K_d}]^\mathrm{T}$ are calculated by a convolutional layer $\mathrm{\Psi}_\beta$ which takes features $\mathcal{H}(\mathrm{W}_i)$, coefficients $\alpha_i$ and a noise vector $z \sim \mathrm{N}(0, \mathrm{I})$ as inputs:
\begin{eqnarray}
  \beta_i = \mathrm{\Psi}_\beta(\mathcal{H}(\mathrm{W}_i), \alpha_i, z).
\end{eqnarray}
By sampling $N_m$ noise vectors $\{z^1,\cdots,z^{N_m}\}$ from the isotropic Gaussian with dimension $dim_z$ and passing them to $\mathrm{\Psi}_\beta$, we have a collection $\{\beta_i^1,\cdots,\beta_i^{N_m}\}$. Next, the basis and multiple deformations are produced by two following convolutional layers $\mathrm{\Psi}_B$ and $\mathrm{\Psi}_D$:
\begin{eqnarray}
\mathrm{S}_{i}^B=\mathrm{\Psi}_B(\alpha_i), \quad
\mathrm{S}_{i}^{D,m}=\mathrm{\Psi}_D(\beta_i^m).
\end{eqnarray}
Note that the atoms $\mathrm{B}_k$ and $\mathrm{D}_l$ are learned as the parameters of $\mathrm{\Psi}_B$ and $\mathrm{\Psi}_D$. Finally, the multiple hypothesis reconstructions $\{\mathrm{S}_i^1,\cdots,\mathrm{S}_i^{N_m}\}$ are generated by:
\begin{eqnarray}
  \mathrm{S}_{i}^m = \mathrm{S}_{i}^B + \mathrm{S}_{i}^{D,m},
\end{eqnarray}
where $m=1\cdots N_m$.

For camera $\mathrm{R}_i$ estimation, we avoid the decomposition ambiguity by simply estimating one rotation matrix for all hypothesis. We follow \cite{Novotny2020C3DPO} to predict the rotation matrix using Rodrigues' Rotation Formula, which is parameterized by a 3 dimensional output of a convolutional layer built upon the extracted features. Given $\mathrm{\Pi}$, we obtain $\mathrm{M}_i=\mathrm{\Pi}\mathrm{R}_i$.

\subsection{Optimizing with Intermediate Loss and Best Hypothesis Selection}
\label{sec:optimization}

We introduce two effective methods for optimizing MHR-Net on the data term $\mathcal{L}_{\mathrm{data}}$.

\noindent
$\textbf{Intermediate Loss.}$ First, we consider the optimization of a single hypothesis $S_i^m$ produced by MHR-Net from $\mathrm{W}_i$. The proposed basis-deformation expression of shapes in Eq.~\ref{eq:shape_expression} is the summation of two linear subspaces. In practice, MHR-Net is prone to using only one flexible subspace $\mathrm{S}_i^D$ for shape expression if we adopt the standard reprojection loss $\left\| \mathrm{W}_i - \mathrm{M}_i\mathrm{S}_i^m \right\|$. To make the basis-deformation expression work as expected, we propose to add an extra reprojection loss to the intermediate results of reconstruction, i.e. the basis $\mathrm{S}_i^B$. This enforces MHR-Net to produce a low-rank and coarse prediction $\mathrm{S}_i^B$ of the 3D shape, thus letting $\mathrm{S}_i^D$ focus on the small and diverse residuals. 
The extra intermediate loss is written as $\left\| \mathrm{W}_i - \mathrm{M}_i\mathrm{S}_i^B \right\|$. 

\noindent
$\textbf{Best Hypothesis Selection.}$ Now we consider the optimization of all hypotheses. There are several approaches to train a model with multiple predictions, including minimizing losses of all predictions, selecting one hypothesis with mixture density~\cite{LiL19}, \emph{etc}. In MHR-Net, we choose to use a pseudo ``hindsight'' loss~\cite{DBLP:conf/nips/LeePCRCB16,DBLP:conf/iccv/RupprechtLDB17}. The original ``hindsight'' loss choose a prediction that is the closest to the ground truth and then calculate the standard single-prediction loss. As in NRSfM we do not have any 3D ground truth, we heuristically use the reprojection error as the criterion to select the best hypothesis among our predictions. Together with the intermediate loss, $\mathcal{L}_{\mathrm{data}}$ is written as:
\begin{eqnarray}
  \mathcal{L}_{\mathrm{data}} = \lambda_B\left\| \mathrm{W}_i - \mathrm{M}_i\mathrm{S}_i^B \right\| + \lambda_F\mathop{\min}_{m}\left\| \mathrm{W}_i - \mathrm{M}_i\mathrm{S}_i^m \right\|,
\label{eq:loss_data}
\end{eqnarray}
where $\lambda_B, \lambda_F$ are balancing factors and $\lambda_B+\lambda_F=1$. Compared to other multi-prediction training approaches, we find that the proposed strategy brings a better reconstruction accuracy to all hypotheses. Moreover, when inferring a single best reconstruction $\mathrm{S}_i^*$ of the input, $\mathrm{S}_i^*$ is naturally obtained with the same best hypothesis strategy.


\noindent
$\textbf{Discussion.}$ (1) The insight of successful hypothesis generation in MHR-Net is to constrain the norm of the flexible deformation subspace. Although the proposed shape representation (Eq.~\ref{eq:shape_expression}) has a large degree of freedom ($\mathrm{S}_i^{D}$ could cause a maximum of $2^{9K_d}$ local minimas~\cite{AkhterSK09}), Eq.~\ref{eq:loss_data} implicitly constrains the norm of $\mathrm{S}_i^{D}$ to be relatively small compared to $\mathrm{S}_i^{B}$ when the balancing factors are chosen as $\lambda_B = 0.8, \lambda_F = 0.2$. On the other hand, one can choose a smaller $\lambda_B$ and adding a diversity loss as in~\cite{LiL20} for generating more diverse hypotheses.

\noindent (2) Moreover, we found that the model with DBSD has a slightly higher $\mathcal{L}_{data}$ and a lower $\mathcal{L}_{reg}$ compared to the deterministic model, which indicates that DBSD leads to a better regularized model. The diverse hypotheses reduce overfitting of MHR-Net and enhance the generalization capability.


\subsection{Procrustean Residual Loss for Regularization}
\label{sec:reg}
In this section, we introduce a novel Procrustean Residual Loss for regularizing the non-rigid shape reconstruction.

\noindent\textbf{Motivation.} Reducing the rigid motions between reconstructed shapes is one of the keys to successful NRSfM. In the previous work of Novotny et al.~\cite{Novotny2020C3DPO}, the Transversal property is proposed to characterize a space where a shape is enforced to appears in a canonical view, such that the effects of rigid motions between same shapes are removed. Implemented with an auxiliary neural network, the Transversal property is effective in performing non-rigid reconstructions.

However, the Transversal property is still restricted to aligning only identical shapes. That means shapes with small differences are not guaranteed to be aligned in a Transversal set. In other words, the effect of a rigid motion is not removed for similar (but not identical) shapes. As regularization on similar shapes is shown to be useful in a recent work \cite{Zeng21}, we are motivated to reduce the rigid motion between similar shapes.

To achieve this objective, we first define two distance measures:
\begin{definition}
Given two non-degenerated shapes $S_i, S_j \in \mathbb{R}^{3 \times N_p}$ and the optimal rotation $R^*$ aligning $S_i$ to $S_j$, the Original distance $\delta_{ori}$ is $\left\| S_i - S_j \right\|_\mathrm{F}$, and the Procrustean distance $\delta_{pro}$ is $\left\| R^*S_i - S_j \right\|_\mathrm{F}$.
\end{definition}
Here, a shape is non-degenerated if $\mathrm{rank}(\mathrm{S}_i)=3$, $\left\| \cdot \right\|_\mathrm{F}$ denotes the Frobenius norm, and the optimal rotation $\mathrm{R}^*$ is obtained with the orthogonal Procrustes~\cite{Schonemann66}.

Next, we use the two defined distances to: (1) test whether $\mathrm{S}_i$ and $\mathrm{S}_j$ are similar or not; (2) if $\mathrm{S}_i, \mathrm{S}_j$ are similar, measure the effect of rigid motions. In step (1), $\mathrm{S}_i, \mathrm{S}_j$ are considered to be similar if $\delta_{pro}<\epsilon$, where $\epsilon$ is a hyper-parameter of similarity threshold. We use the Procrustean distance in this step since it is agnostic of rigid rotations. In step (2), we propose to measure the effect of rigid motions with the \emph{Procrustean Residual} $\delta_{res}$, calculated as:
\begin{eqnarray}
  \delta_{res} = \delta_{ori} - \delta_{pro}.
\end{eqnarray}
The Procrustean Residual indicates how much the Original distance can be reduced with rigid motions, and $\delta_{res}$ reaches zero if and only if $\mathrm{S}_i$ and $\mathrm{S}_j$ are already optimally aligned (i.e. $\mathrm{R}^*=\mathrm{I}$). Therefore, the undesirable rigid motion between $\mathrm{S}_i,\mathrm{S}_j$ is reduced when we minimize $\delta_{res}$.

\noindent\textbf{Loss design.} We now introduce the Procrustean Residual Loss for NRSfM regularization. This loss function realizes the minimization\footnote{Note that in general $\delta_{res}$ cannot be reduce to exactly zero for all pairs simultaneously.} of $\delta_{res}$ and is developed in a differentiable way.

Given two shapes $\mathrm{S}_i, \mathrm{S}_j$ randomly sampled from the network prediction batch, the optimal rotation that aligns $\mathrm{S}_i$ to $\mathrm{S}_j$ is calculated with the orthogonal Procrustes~\cite{Schonemann66}:
\begin{eqnarray}
  \mathrm{R}^*_{i,j}=\mathrm{VU^T}, \quad \mathrm{S}_i\mathrm{S}_j^\mathrm{T}=\mathrm{U\Sigma V^T},
\label{eq:ortho_procrucst}
\end{eqnarray}
where $\mathrm{U\Sigma V^T}$ is the singular value decomposition of $\mathrm{S}_i\mathrm{S}_j^\mathrm{T}$.

Next, we calculate the (normalized) Procrustean distance and the Procrustean Residual as follows:
\begin{eqnarray}
  \bar{\delta}_\mathrm{pro} = \frac{\left\| \mathrm{R}^*_{i,j}\mathrm{S}_i - \mathrm{S}_j \right\|_\mathrm{F}}{\left\| \mathrm{S}_j \right\|_\mathrm{F}}, \quad \bar{\delta}_\mathrm{res} = \frac{\left\| \mathrm{S}_i - \mathrm{S}_j \right\|_\mathrm{F}}{\left\| \mathrm{S}_j \right\|_\mathrm{F}} - \bar{\delta}_\mathrm{pro}.
\end{eqnarray}
We normalize the differences with $\left\| \mathrm{S}_j \right\|_\mathrm{F}$ to make the loss numerically stable. The Procrustean Residual Loss is:
\begin{eqnarray}
\mathcal{L}_{\mathrm{res}}(\mathrm{S}_i,\mathrm{S}_j) = \mathop{\rho}(\bar{\delta}_{\mathrm{pro}},\epsilon)\cdot \bar{\delta}_\mathrm{res},
\end{eqnarray}
where $\mathop{\rho}(\bar{\delta}_{\mathrm{pro}},\epsilon) = 1$ if $\bar{\delta}_{\mathrm{pro}} < \epsilon$, else $\mathop{\rho}=0$.

The practical problem of the proposed loss function is that $\mathcal{L}_{\mathrm{res}}$ contains a non-differentiable operation SVD in Eq.~\ref{eq:ortho_procrucst}. 
To make $\mathcal{L}_{\mathrm{res}}$ differentiable, one effective way is to use the Lagrange multiplier method on Lie Group~\cite{DBLP:journals/tip/ParkLK18,DBLP:conf/eccv/ParkLK20} for a closed-form partial derivative, and another way is to leverage modern auto-grad libraries where the numeric computation of SVD is differentiable~\cite{9099404,WangL21,Zeng21}. In this paper, we choose to use a simple alternative approach by detaching $\mathrm{R}^*_{i,j}$ from the computation graph, namely $\mathrm{R}^*_{i,j}$ is viewed as a constant matrix. In such way, $\mathcal{L}_{\mathrm{res}}$ only involves standard differentiable operations of inputs $\mathrm{S}_i,\mathrm{S}_j$, which frees us from the calculation of SVD gradient and keeps the model computationally efficient.

For the regularization term $\mathcal{L}_{\mathrm{reg}}$ of MHR-Net, we apply a canonicalization loss $\mathcal{L}_{\mathrm{cano}}$\footnote{Please refer to the supplementary material or~\cite{Novotny2020C3DPO} for details.}~\cite{Novotny2020C3DPO} to the deterministic basis $\mathrm{S}_i^B$ and $\mathcal{L}_{\mathrm{res}}$ to $\mathrm{S}_i^*$, leading to:
\begin{eqnarray}
\mathcal{L}_{\mathrm{reg}} = \mathcal{L}_{\mathrm{cano}} + \lambda_{\mathrm{res}}\mathcal{L}_{\mathrm{res}},
\end{eqnarray}
where $\lambda_{\mathrm{res}}$ is the weight of Procrustean Residual Loss. Although only including $\mathcal{L}_{\mathrm{res}}$ in $\mathcal{L}_{\mathrm{reg}}$ is possible and produces good results, we empirically find that using two losses jointly leads to a better performance.

\section{Experiments}
We evaluate the proposed MHR-Net in two aspects: (1) The reconstruction accuracy of the best hypothesis. (2) The multiple hypothesis reconstructions. We also make an in-depth analysis of proposed components.

\subsection{Datasets and Experimental Setups}
\noindent
\textbf{Human3.6M}~\cite{h36m_pami}. It is the largest 3D human pose dataset with a total of 3.6 million frames. It contains 15 different activities performed by 7 professional actors and captured by four cameras.  We follow the common protocols to use five subjects (S1, S5-8) as the training set and two subjects (S9 and S11) as the testing set. We adopt the widely-used pre-processing from Pavllo et al.~\cite{DBLP:conf/cvpr/PavlloFGA19}.

\noindent
\textbf{300VW}~\cite{DBLP:conf/iccvw/ShenZCKTP15}. The 300VW is a large-scale facial landmark dataset. It has a total of 114 sequences with 2D annotations of 68 landmarks. Following \cite{DBLP:conf/eccv/ParkLK20}, we use the subset of 64 sequences from 300VW, and divide them into a training set and a testing set of 32 sequences respectively. As 3D ground-truth is not provided, we follow \cite{DBLP:conf/eccv/ParkLK20} to adopt the results from \cite{DBLP:conf/iccv/BulatT17} as 3D ground-truths.

\noindent
\textbf{SURREAL}~\cite{DBLP:conf/cvpr/Varol0MMBLS17}. The SURREAL dataset contains 6 million synthetic human images with large variations in shape, view-point and pose. The 6,890 dense 3D points are obtained by fitting SMPL~\cite{DBLP:journals/tog/LoperM0PB15} to CMU MOCAP dataset. Following~\cite{DBLP:conf/eccv/ParkLK20,WangL21}, the training and testing sets include 5,000 and 2,401 frames selected from the full dataset, respectively. 

\noindent
\textbf{Metrics}. We adopt the following two metrics: \\
(1) MPJPE: It stands for the mean per joint error, which is calculated as $\frac{1}{N_p}\left\| \mathrm{S}_i - \mathrm{S}_i^{\mathrm{gt}} \right\|_1$. To address the reflection ambiguity, we follow \cite{Novotny2020C3DPO,DBLP:conf/eccv/ParkLK20,WangL21} to report the minimal error with ground-truth between original and flipped shapes.
\\
(2) Normalized Error (NE): It reflects the relative estimation error and is computed by: $\frac{\left\| \mathrm{S}_i - \mathrm{S}_i^{\mathrm{gt}} \right\|_\mathrm{F}}{ \left\|\mathrm{S_{gt}} \right\|_\mathrm{F}}$.



\begin{table}[t]
\footnotesize
\small
\resizebox{\textwidth}{!}{
\begin{tabular}{l|lllllllllllllll|l} \toprule
Methods (Ortho.) & Direct. & Discuss & Eating & Greet & Phone & Pose & Purch. & Sitting & SittingD. & Smoke & Photo & Wait & Walk & WalkD. & WalkT. & Mean  \\ \hline
CSF2~\cite{DBLP:conf/cvpr/GotardoM11}     & 87.2       &  90.1      &  96.1      &  95.9    &  102.9     &  92.1      & 99.3      & 129.8        &  136.7     &  99.5     & 120.1    &  95.2   &  90.8      &  102.4    &  89.2     & 101.6 \\
SPM~\cite{DaiLH14}    & 65.3       & 68.7       &  82.0      & 70.1     &  95.3    & 65.1    &  71.9    & 117.0        & 136.0    & 84.3    & 88.9    & 71.2   & 59.5     & 73.3    & 68.3    & 82.3 \\
C3DPO~\cite{Novotny2020C3DPO}  & \textbf{56.1}       & 55.6     & 62.2    & \textbf{66.4}    & 63.2    & \textbf{62.0}     & 62.9     & 76.3       & 85.8    & \textbf{59.9}    & 88.7   & 63.3   & 71.1    & 70.7    & 72.3    & 67.8 \\
PRN~\cite{DBLP:conf/eccv/ParkLK20}    & 65.3      & 58.2      & 60.5     & 73.8    & \textbf{60.7}    & 71.5     & 64.6      & 79.8        & 90.2     & 60.3    & \textbf{81.2}    & 67.1   & \textbf{54.4}     & \textbf{61.2}    &  \textbf{65.6}    & 66.7 \\ \hline
MHR-Net (Ours)     & 60.3       & \textbf{54.3}       & \textbf{55.5}      & 67.9     & 67.7     & 69.5      & \textbf{61.3}       & \textbf{69.7}         & \textbf{83.2}     & 67.6     & 85.3    & \textbf{61.7}    & 61.9      & 63.4     & 68.2     & \textbf{65.8} \\ \hline \hline
Methods (Persp.) & Direct. & Discuss & Eating & Greet & Phone & Pose & Purch. & Sitting & SittingD. & Smoke & Photo & Wait & Walk & WalkD. & WalkT. & Mean  \\ \hline
PoseGAN~\cite{DBLP:journals/corr/abs-1803-08244}    & -       & -       & -      & -     & -     & -      & -       & -         & -     & -     & -    & -    & -      & -     & -     & 130.9 \\
SFC~\cite{DBLP:conf/3dim/KongZKL16} & -       & -       & -      & -     & -     & -      & -       & -         & -     & -     & -    & -    & -      & -     & -     & 218.0 \\
Consensus~\cite{8778692} & -       & -       & -      & -     & -     & -      & -       & -         & -     & -     & -    & -    & -      & -     & -     & 120.1 \\
DNRSFM~\cite{9099404}    & -       & -       & -      & -     & -     & -      & -       & -         & -     & -     & -    & -    & -      & -     & -     & 101.6 \\
Wang et al.~\cite{WangKL19}  & -       & -       & -      & -     & -     & -      & -       & -         & -     & -     & -    & -    & -      & -     & -     & 86.2 \\
C3DPO~\cite{Novotny2020C3DPO}   & 96.8    & 85.7    & 85.8   & 107.1     & 86.0   & 96.8  & 93.9   & 94.9    & 96.7   & 86.0   & 124.3 & 90.7 & 95.2 & 93.4   & 101.3  & 95.6 \\
PRN~\cite{DBLP:conf/eccv/ParkLK20}     & 93.1    & 83.3    & 76.2   & 98.6  & 78.8  & 91.7   & 81.4   & 87.4    & 91.6      & 78.2  & 104.3 & 89.6 & 83.0   & 80.5   & 95.3   & 86.4 \\
PAUL~\cite{WangL21}    & -       & -       & -      & -     & -     & -      & -       & -         & -     & -     & -    & -    & -      & -      & -     & 88.3 \\
PoseDict~\cite{XuCL021} & 74.6 & 82.9 & 77.0 & 86.7 & 80.0 & 79.2 & 94.2 & 88.4 & 124.0 & 77.1 & 103.8 & 80.8 & 78.8 & 94.2  & 78.3  & 85.6  \\
ITES~\cite{XuCL021}    &  77.6  & 77.3 & 77.1 & 77.3 & 77.3 & 77.4  & 77.3 & 77.2 & \textbf{77.3} & 77.1 & \textbf{77.1} & 77.5 & 77.3 & \textbf{77.2} & 77.5 & 77.2 \\ \hline
MHR-Net (Ours)     &  \textbf{62.8}  &  \textbf{68.3}  & \textbf{62.2} &  \textbf{73.9}  & \textbf{73.7}  &  \textbf{67.0}  &  \textbf{70.2} &  \textbf{76.7}  & 100.0 & \textbf{71.5} & 90.0 & \textbf{72.3} &  \textbf{68.8}  & 80.2 & \textbf{71.0} & \textbf{72.6} \\ 
\bottomrule
\end{tabular}}
\caption{Quantitative results on Human3.6M Dataset.}
\label{tab:results_h36m}
\end{table}

\subsection{Main Results}
In this subsection and Sec.~\ref{sec:ablation}, we treating MHR-Net as a single-prediction model by using the best hypothesis $\mathrm{S}_i^*$. We report the standard NRSfM evaluation results of MHR-Net on three datasets. 

For Human3.6M, we test the performance of MHR-Net under two settings: orthographic camera and perspective camera. The major competitors of MHR-Net are state-of-the-art deep NRSfM models, including C3DPO~\cite{Novotny2020C3DPO}, DNRSFM~\cite{9099404}, PRN~\cite{DBLP:conf/eccv/ParkLK20}, PAUL~\cite{WangL21}, ITES~\cite{XuCL021}. In Table.~\ref{tab:results_h36m}, we report the MPJPE of all frames and 15 individual activities on the test set. We also includes classic methods like Consensus~\cite{8778692}, SFC~\cite{DBLP:conf/3dim/KongZKL16} for comparison. As shown in Tab.~\ref{tab:results_h36m}, MHR-Net outperforms all competing methods overall in both orthographic and perspective experiments. These results on the challenging Human3.6M dataset verify the effectiveness of MHR-Net on reconstructing highly-flexible human poses.

For 300VW dataset and SURREAL dataset, we compare MHR-NET with C3DPO~\cite{Novotny2020C3DPO}, PRN~\cite{DBLP:conf/eccv/ParkLK20}, PR~\cite{DBLP:journals/tip/ParkLK18} and PAUL~\cite{WangL21}. The Normalized Error results are shown in Tab.~\ref{tab:results_300vw} and Tab.~\ref{tab:results_surreal} respectively. These outcomes validate that MHR-Net is capable to perform accurate reconstruction of both facial landmarks and dense meshes. It is worth noting that MHR-Net recovers the dense point clouds of SURREAL dataset without splitting it into several subsets, unlike \cite{DBLP:conf/eccv/ParkLK20}. This is achieved by avoiding the burdensome SVD of matrices whose scales are related to $N_p$. With the differentiable design, MHR-Net shares the same level of scalability as SVD-free methods while achieving better performance.

\begin{table}[!tb]
    \footnotesize
    \centering  
    \begin{minipage}{.55\linewidth}
      \centering
        \begin{tabular}[t]{lc}
        \toprule
        Model & NE\\
        \midrule
        C3DPO~\cite{Novotny2020C3DPO} & 0.3509\\
        PRN~\cite{DBLP:conf/eccv/ParkLK20} & 0.1377 \\
        PAUL~\cite{WangL21} & 0.1236 \\
        \midrule
        MHR-Net (w/o $\mathcal{L}_{\mathrm{res}}$) & 0.1388 \\
        MHR-Net & \textbf{0.1192} \\
        \bottomrule
        \end{tabular}
        \caption{Results on SURREAL.}
        \label{tab:results_surreal}
    \end{minipage}%
    \begin{minipage}{.45\linewidth}
      \centering
        \begin{tabular}[t]{lc}
        \toprule
        Model & NE\\
        \midrule
        CSF2~\cite{DBLP:conf/cvpr/GotardoM11} & 0.2751\\
        PR~\cite{DBLP:journals/tip/ParkLK18} & 0.2730 \\
        C3DPO~\cite{Novotny2020C3DPO} & 0.1715 \\
        PRN~\cite{DBLP:conf/eccv/ParkLK20} & 0.1512 \\
        \midrule
        MHR-Net & \textbf{0.1007} \\
        \bottomrule
        \end{tabular}
        \caption{Results on 300VW.}
        \label{tab:results_300vw}
    \end{minipage}
\end{table}

\subsection{Ablation Study}
\label{sec:ablation}

We show the effectiveness of the important designs in MHR-Net. We conduct the experiments on perspective Human 3.6M dataset. BD and IL in Tab.~\ref{tab:ablation} are short for Basis-Deformation (Eq.~\ref{eq:shape_expression}) and Intermediate Loss (Sec.~\ref{sec:optimization}).

\noindent
\textbf{Basis-Deformation and Intermediate Loss.} 
We setup several ablated models: (1) Baseline: We use a modified PoseDict~\cite{XuCL021} as baseline. We replace the invariance loss of PoseDict with the canonicalization loss~\cite{Novotny2020C3DPO}, and it works slightly better than PoseDict. In Baseline, only the basis and camera are estimated. (2) Baseline with Basis-Deformation. The deformation here is implemented deterministically. (3) Baseline with Basis-Deformation and Intermediate Loss. Comparing the results of (1) and (2) in Tab.~\ref{tab:ablation}, we observe the degradation of performance. This implies that a naive extension of the deformation subspace will harm the regularization of low-rankness and lead to the failure of non-rigid reconstruction. By adding an intermediate loss in (3), the MPJPE is greatly reduced from 83.5 to 75.5, which is already better than the MPJPE (77.2) of the most competitive method ITES.

\begin{table}[t]
\small
\centering
\begin{tabular}{c|>{\centering}p{1.00cm}>{\centering}p{1.00cm}>{\centering}p{1.00cm}>{\centering}p{1.00cm}>{\centering}p{1.00cm} | c}
\toprule
No.  & BD & IL & DBSD & Optim. & $\mathcal{L}_{\mathrm{res}}$ & MPJPE \\ \hline
1    & \xmark  & \xmark & \xmark & \xmark & \xmark & 83.5 \\
2    & \cmark  &        &        &        &        & 198.3  \\
3    & \cmark  & \cmark &        &        &        & 75.5 \\
4    & \cmark  & \cmark & \cmark & Best   &        & 73.7 \\
5    & \cmark  & \cmark & \cmark & Worst  &        & 78.4 \\
6    & \cmark  & \cmark & \cmark & MD     &        & 77.9 \\
7    & \cmark  & \cmark & \cmark & Worst  & \cmark & 76.2 \\
8    & \cmark  & \cmark & \cmark & MD     & \cmark & 77.3 \\
9   & \cmark  & \cmark & \cmark & Best   & \cmark & 72.6 \\ \bottomrule
\end{tabular}
\caption{Ablation study results.}
\label{tab:ablation}
\end{table}

\noindent
\textbf{Stochastic Deformation and Hypothesis Optimization Strategy.} We now use the deterministic basis and stochastic deformation with best hypothesis selection strategy, indicated by (4) in Tab.~\ref{tab:ablation}. We compare the proposed design with two alternatives: (5) Worst Hypothesis. In this strategy, we choose to optimize the hypothesis with the largest re-projection error, which is the opposite of the best hypothesis strategy. The intuition of Worst Hypothesis is that it tries to minimize the upper bound of errors. (6) Mixture Density (MD)~\cite{LiL19}, where a hypothesis is selected by sampling from a learned mixture density. As we do not have ground-truth labels for training, we use the uniform categorical distribution instead of a learned distribution. The results in Tab.~\ref{tab:ablation} demonstrates that the combination of DBSD and the Best Hypothesis strategy produce the best performance.

\noindent
\textbf{Procrustean Residual Loss.} We show the effectiveness of the Procrustean Residual Loss $\mathcal{L}_{\mathrm{res}}$. As reported in Tab.~\ref{tab:ablation}, the full model (9) has a better MPJPE of 72.6. Despite the fact that the previous state-of-the-art method ITES uses two networks and the prior knowledge of human pose, MHR-Net outperforms ITES by 4.6 MPJPE. By adding $\mathcal{L}_{\mathrm{res}}$ to models with alternative MD and Worst Hypothesis, these two models (7) and (8) also outperform the corresponding models (5) and (6) without $\mathcal{L}_{\mathrm{res}}$. 
Moreover, the improvement by using $\mathcal{L}_{\mathrm{res}}$ is also significant on the dense mesh dataset SURREAL, as indicated in Tab.~\ref{tab:results_surreal}.

\begin{figure}[t]
\begin{center}
\includegraphics[width=1.0\linewidth]{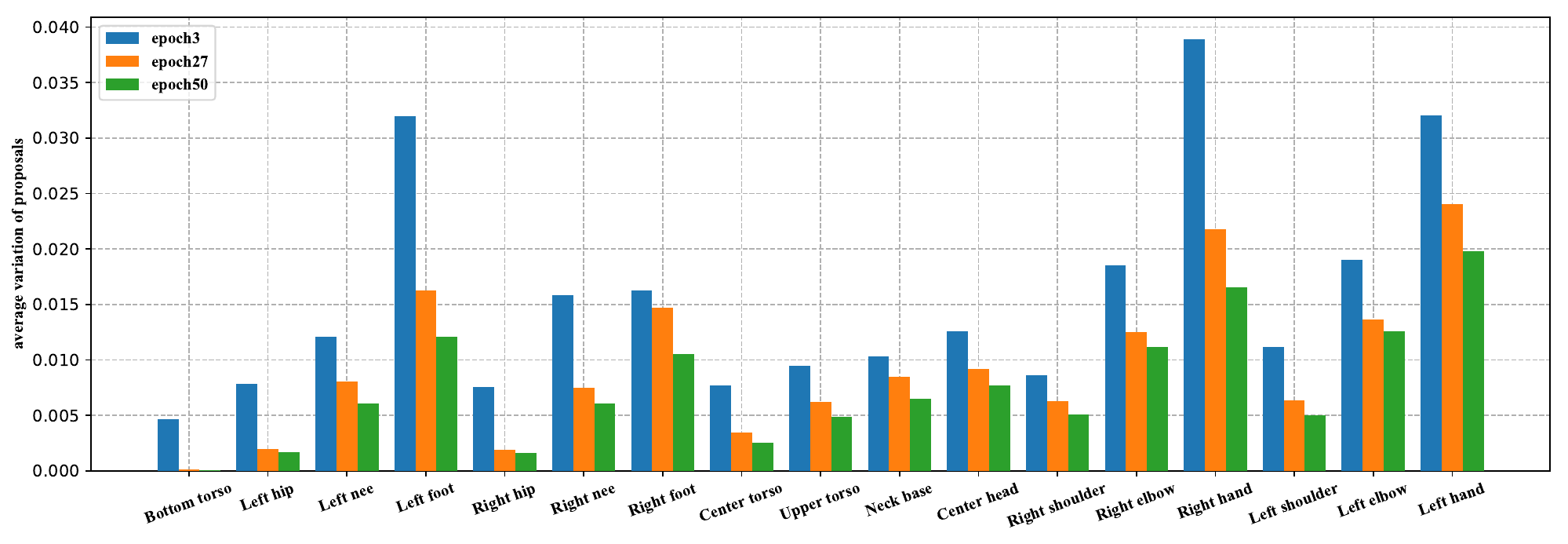}
\end{center}
\caption{Point-wise variation of deformations.}
\label{fig:var_deform}
\end{figure}

\subsection{Analysis of Stochastic Deformation}

\noindent
\textbf{Point-wise variation of deformations.} We measure the variation of stochastic deformations of point $j$ as $\frac{1}{N_f}\sum_{i=1}^{N_f}\mathop{\max}_{m,n}\left\| \mathrm{S}_{i,j}^{D,m} - \mathrm{S}_{i,j}^{D,n} \right\|_\mathrm{F}$, where the subscript $j$ denotes the $j$-th point. As shown in Fig.~\ref{fig:var_deform}, the variation of deformations decreases as the training processes. One tentative interpretation is that MHR-Net searches solutions with a more diverse set of hypotheses in the early stage of training, and produces less diverse (or more confident) hypotheses after convergence. Moreover, the variation also differs between points. We observe the largest variations at \texttt{Left/Right Hand/Elbow/Foot}, which is consistent with the common sense that these are most flexible parts of body.

\noindent
\textbf{Accuracy of other hypotheses}. To verify the accuracy of non-best hypotheses, we evaluate the MPJPE of the worst (largest reprojection error) hypothesis on Human3.6M. Compared to the best hypothesis, the results of worst hypothesis show a decline of -0.5 and -0.1 MPJPE at epoch 10 and epoch 50 with $N_m=50$. This verifies that other hypotheses from MHR-Net are also accurate.

\begin{figure}[t]
\begin{center}
\includegraphics[width=0.50\linewidth]{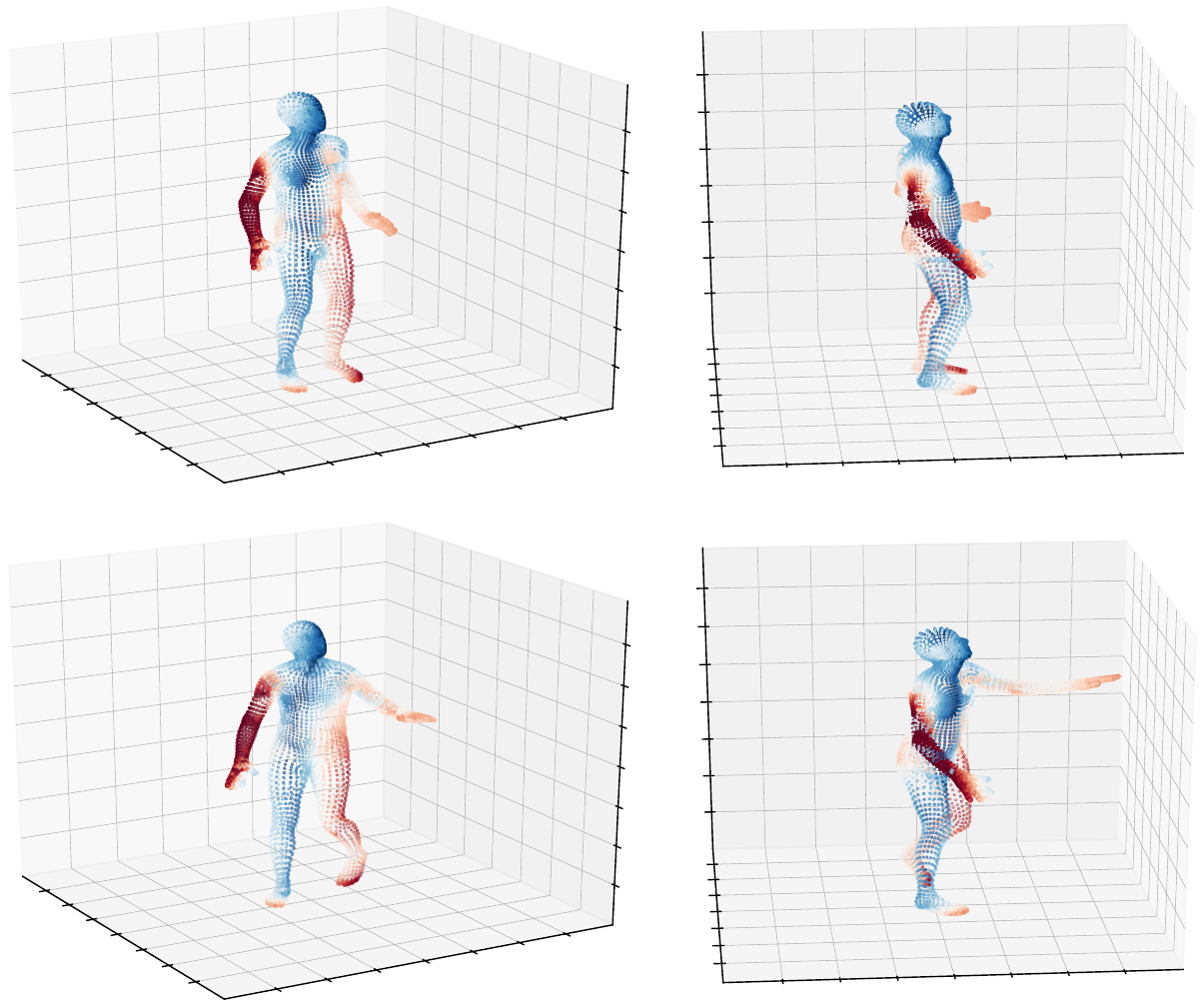}
\end{center}
\caption{Visualization of reconstructions on SURREAL. Points with large deformations are marked in red. Best viewed in color.}
\label{fig:vis_surreal} 
\end{figure}

\noindent
\textbf{Visualizing largely-deformed regions of meshes.} Another advantage of MHR-Net is that we can use $\left\| \mathrm{S}_i^D \right\|$ as an indicator of the degree of deformations. We visualize the largely-deformed regions of reconstructed dense point clouds from SURREAL dataset, as shown in Fig.~\ref{fig:vis_surreal}. The visualization clearly illustrates the largely deformed parts of the body, which is helpful for better understanding of non-rigid reconstruction.


\begin{figure}[t]
\begin{center}
\includegraphics[width=1.0\linewidth]{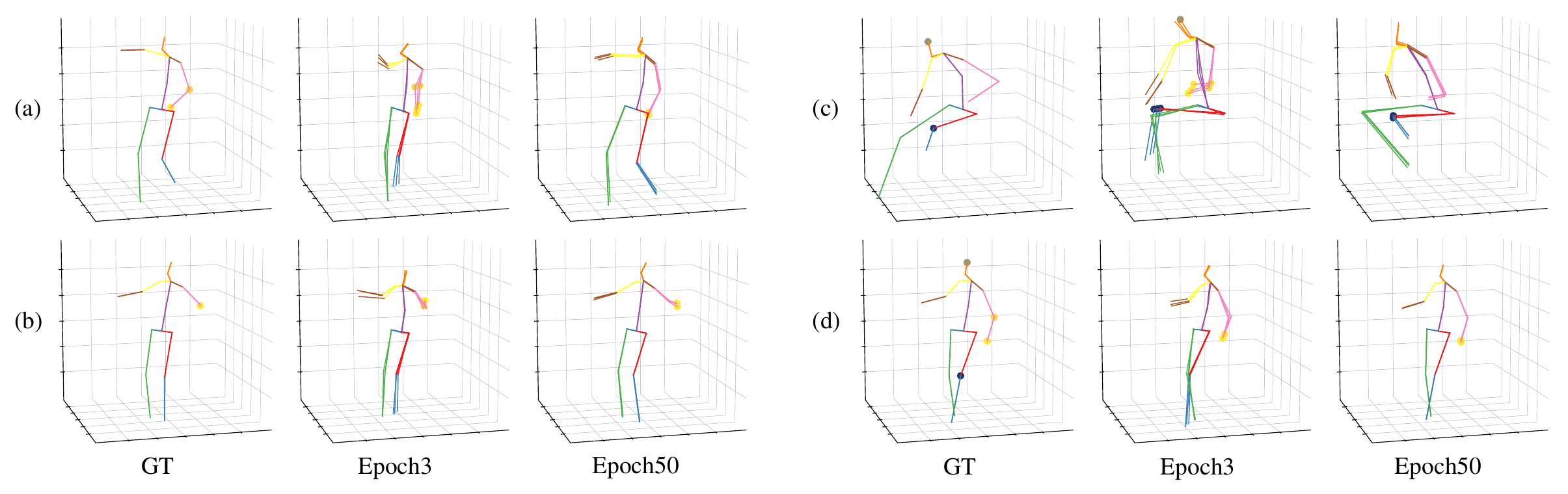}
\end{center}
\caption{Visualization of hypotheses ($N_m=3$) on Human3.6M test set.}
\label{fig:vis_hypo}
\end{figure}

\section{Conclusion}
We propose MHR-Net, a novel approach for reconstructing non-rigid shapes from 2D observations. To the best of our knowledge, MHR-Net is the first method produces  multiple high-quality hypotheses for NRSfM. With the proposed DBSD scheme and optimization strategy, MHR-Net is capable to generate multiple reconstructions and achieves state-of-the-art shape recovery accuracy. Furthermore, we introduce a Procrustean Residual Loss for enhancing performance.

\section*{Acknowledgement}
This work was partially done when Haitian Zeng interned at Baidu Research. This work was partially supported by ARC DP200100938. We thank Dr. Sungheon Park for sharing the SURREAL dataset. We thank all reviewers and area chairs for their valuable feedback. 



%
%

\bibliographystyle{splncs04}
\end{document}